\pdfoutput=1

\documentclass[11pt]{article}

\usepackage[preprint]{acl}
\usepackage{math}

\usepackage{times}
\usepackage{latexsym}

\usepackage[T1]{fontenc}

\usepackage[utf8]{inputenc}

\usepackage{microtype}

\usepackage{inconsolata}

\title{To be Continuous, or to be Discrete, Those are \emph{Bits} of Questions}

\author{Yiran Wang\quad Masao Utiyama \\
        National Institute of Information and Communications Technology (NICT) \\
        \texttt{yiran.wang@nict.go.jp\quad mutiyama@nict.go.jp}} 

\begin{document}
\maketitle
\begin{abstract}
Recently, binary representation has been proposed as a novel representation that lies between continuous and discrete representations. It exhibits considerable information-preserving capability when being used to replace continuous input vectors. In this paper, we investigate the feasibility of further introducing it to the output side, aiming to allow models to output binary labels instead. To preserve the structural information on the output side along with label information, we extend the previous contrastive hashing method as structured contrastive hashing. More specifically, we upgrade CKY from label-level to bit-level, define a new similarity function with span marginal probabilities, and introduce a novel contrastive loss function with a carefully designed instance selection strategy. Our model\footnote{\url{https://github.com/speedcell4/parserker}} achieves competitive performance on various structured prediction tasks, and demonstrates that binary representation can be considered a novel representation that further bridges the gap between the continuous nature of deep learning and the discrete intrinsic property of natural languages.
\end{abstract}
 
\section{Introduction} 

Bridging the gap between the continuous nature of deep learning and the discrete intrinsic property of natural languages has been one of the most fundamental and essential questions since the very beginning. Continuous representation makes the training of neural networks effective and efficient. Nowadays, representing discrete natural languages in continuous format is the first and foremost step to leveraging the capabilities of deep learning. One could even argue that the exhilarating advancements in natural language processing in the past decade can largely be attributed to the word embedding technique, as it is the first successful attempt.\par

\begin{figure}[!t]
  \centering
  \includegraphics[width=\columnwidth]{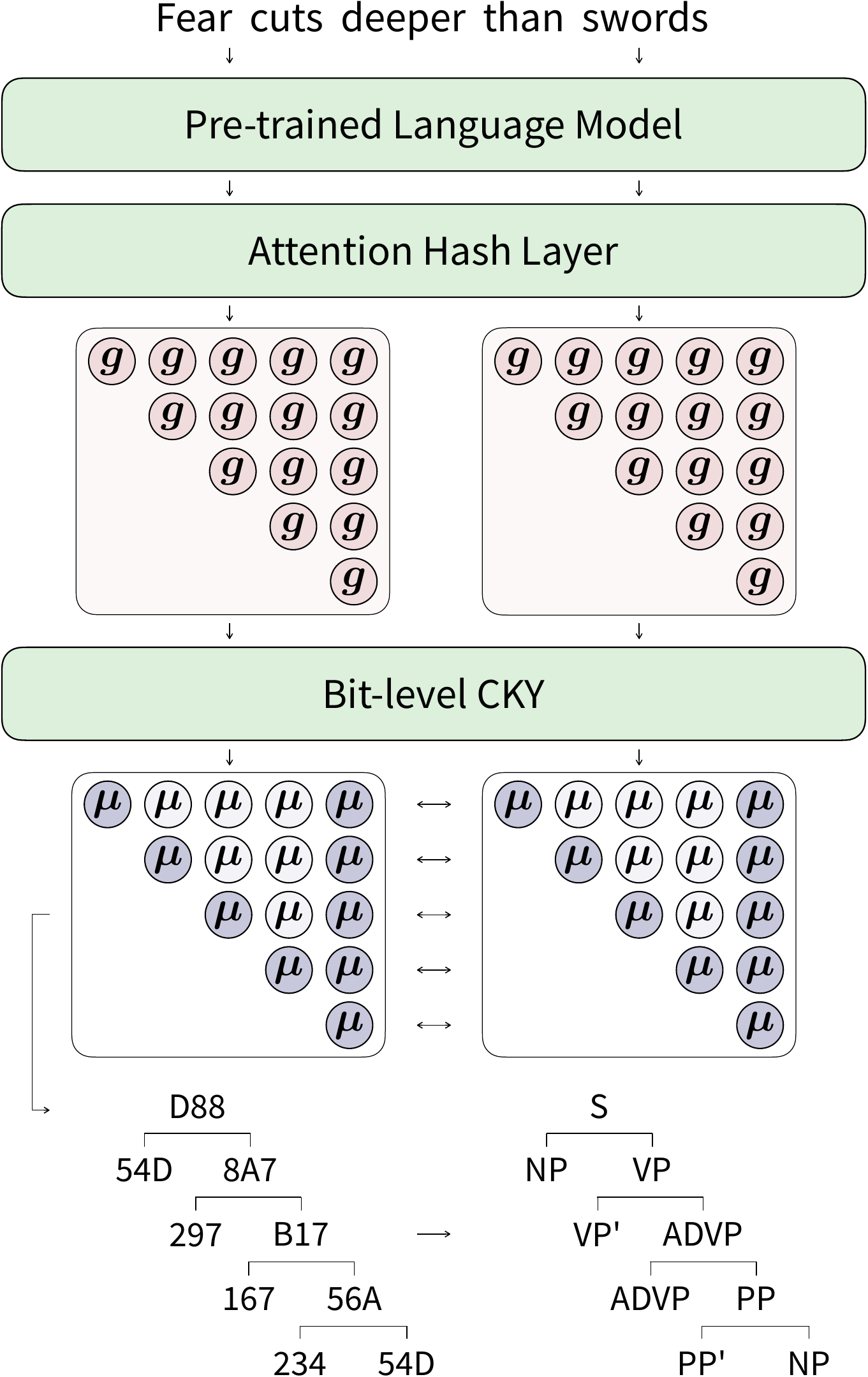}
  \caption{The model architecture. The attention hash layer produces span scores (pink circles), we only use the upper triangular part of these scores and feed them into the bit-level CKY to obtain the marginal probabilities of all valid spans (purple circles). During training, we only select the spans on the target trees for structured contrastive hashing and leave the other spans unused (transparent purple circles). During inference, as shown at the bottom, our model parses sentences by returning trees with label codes (hexadecimal numbers), which are then translated back to the original labels.}
  \label{fig:arch}
\end{figure}

Word embedding \cite{NIPS2000_728f206c, mikolov2013efficient, mikolov2013distributed} technique replaces the vocabulary-sized one-hot word representations with compact continuous vectors. Since then, input and output layers incorporating embedding matrices have become standard components in neural models. Discrete tokens are mapped into continuous vectors by looking up the corresponding index in it, and continuous vectors are mapped back to discrete tokens by searching the most similar one from it. However, the essence of this operation is still one-hot encoding, even the following subword tokenization techniques \cite{sennrich-etal-2016-neural,kudo-2018-subword} attempt to mitigate this issue by decomposing words into subword units, these approaches still require the building vocabularies and embedding matrices that consists of tens of thousands of tokens. In the era of large language models \cite{openai2023gpt4,touvron2023llama}, these embedding matrices typically account for a considerable number of parameters, especially in cross-lingual models. Besides, parameter updates also solely rely on the sparse gradients backpropagated to the limited tokens present in sentences. Moreover, imposing structural constraints on continuous representations to model relations among tokens is considered difficult, whereas it is easy and common in discrete representations. Therefore, further bridging the gap has become increasingly important nowadays.\par

Recently, \citet{wang-etal-2023-24} introduced a novel binary representation that lies between continuous and discrete representations. They proposed a contrastive hashing method to compress continuous hidden states into binary codes. These codes contain all the necessary task-relevant information, and using them as the only inputs can reproduce the performance of the original models. Unlike associating each token with only a single vector, their method allocates multiple bits to each token, and the token representation can be constructed by concatenating these bit vectors. In other words, their binary representation breaks tokens down into combinations of semantic subspaces. As a result, replacing the token embedding matrix in the input layer with a tiny bit embedding matrix without sacrificing performance becomes possible.\par

In this paper, we explore the possibility of further introducing this representation to output layers. In the input layer, structural information can only be implicitly obtained by introducing the task loss as an auxiliary. However, the output layers often involve complex intra-label constraints, especially for structured prediction tasks, structural information can and should be explicitly preserved along with plain label information. Therefore, we attempt to endow models with this capability by extending previous contrastive hashing to structured contrastive hashing.\par

We begin by upgrading the CKY, which parses sentences, returns spans with discrete labels, to support binary format labels (\S\ref{sec:bit_cky}). Subsequently, we define a new similarity function by using span marginal probabilities obtained from this bit-level CKY (\S\ref{sec:similarity}) to jointly learn label and structural information. Furthermore, we conduct a detailed analysis of several widely used contrastive learning losses, identifying the geometric center issue, and introduce a novel contrastive learning loss to remedy it (\S\ref{sec:instance_selection}) through carefully selecting instances. By doing so, we show that it is feasible to introduce binary representation to output layers and have them output binary labels on trees. Moreover, since our model is based on contrastive learning, it also benefits from its remarkable representation learning capability, resulting in better performance than existing models. We conduct experiments on constituency parsing and nested named entity recognition. Experimental results (\S\ref{sec:main_results})  demonstrate that our models achieve competitive performance with only around 12 and 8 bits, respectively.\par

\begin{figure}[!t]
  \centering
  \includegraphics[width=0.5\columnwidth]{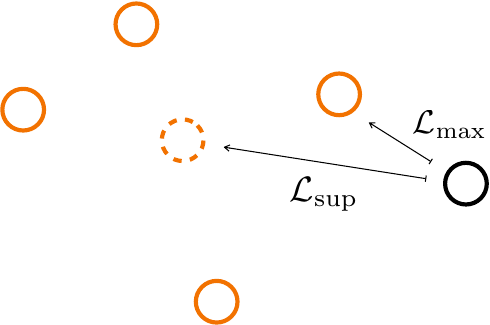}
  \caption{An example of the geometric center issue. Orange circles are positive to the black circle instance, while the dotted orange circle is their geometric center. The difference between $\loss_{\text{sup}}$ and our $\loss_{\text{max}}$ is that we target the closest positive instead of their geometric center.}
  \label{fig:center}
\end{figure}

\section{Background}

\subsection{Constituency Parsing}
\label{sec:parsing}

For a given sentence $w_1,\dots,w_n$, constituency parsing aims at detecting its hierarchical syntactic structures. Previous work \cite{stern-etal-2017-minimal,gaddy-etal-2018-whats,kitaev-klein-2018-constituency,kitaev-etal-2019-multilingual,ijcai2020p0560} decompose tree score $g(\ve{t})$ as the sum of its constituents scores, \begin{gather}
  g(\ve{t}) = \sum_{\langle l_i,r_i,y_i\rangle\in \ve{t}}{g(l_i,r_i,y_i)}
\end{gather} where $l_i$ and $r_i$ indicate the left and right boundary of the $i$-th span, and $y_i\in \ca{y}$ stands for the label. Constituent score $g(l_i,r_i,y_i)$ reflects the joint score of selecting the specified span and assigning it the specified label. Previous work \cite{kitaev-klein-2018-constituency,ijcai2020p0560} commonly compute this score using a linear or bilinear component. Under the framework of graphical probabilistic models, they can efficiently compute the conditional probability by applying the CKY algorithm. \begin{equation}
  p(\ve{t}) = \dfrac{\exp{g(\ve{t})}}{Z\equiv \sum_{\ve{t}'\in\ca{t}}{\exp{g({\ve{t}'})}}} \label{eq:prob}
\end{equation} $Z$ is commonly known as the partition function which enumerates all valid constituency trees.\par

Besides, marginal probability $\mu(l_i,r_i, y_i)$ is also frequently mentioned. It stands for the proportion of scores for all trees that include the specified span with the specified label. As noted by \citet{eisner-2016-inside}, computing the partial derivative of the log partition with respect to the span score is an efficient approach to obtain the marginal probability. \begin{equation}
  \mu(l_i,r_i,y_i) = \dfrac{\partial\,\log{Z}}{\partial\,g(l_i,r_i,y_i)}
\end{equation} 

Intuitively, marginal probability indicates the joint probability of selecting a specified span with a specified label. Therefore, it is easy to notice that merely summing the marginal probabilities for all labels of a given span does not always yield 1, i.e., $\sum_{y'\in\ca{Y}}{\mu(l_i,r_i,y')}\not\equiv 1$, as there is no guarantee that this span will be selected. In other words, marginal probabilities contain not only label information but also structural information. If a span is unlikely to be selected, its marginal probability will not be high regardless of the label.\par

\subsection{Contrastive Hashing}
\label{sec:contrastive}

Contrastive learning \cite{He_2020_CVPR,gao-etal-2021-simcse,NEURIPS2020_d89a66c7} is an effective yet simple representation learning method, which involves pulling together positive pairs and pushing apart negative pairs in a metric space. Recently, \citet{wang-etal-2023-24} extended this approach as contrastive hashing. They append an untrained transformer to the end of a pre-trained language model and use its attention scores for both task learning and hashing. Specifically, its entire attention probabilities $a^k_{i,j}$ are used to compute hidden states for downstream tasks as usual, and its diagonal entries $s^k_{i,i}$ of the attention scores are employed for hashing. \begin{gather}
  s^k_{i,j} = \dfrac{(\ma{w}^Q_k\ve{h}_i)^{\top} (\ma{w}^K_k\ve{h}_j)}{\sqrt{d_k}} \\
  a^k_{i,j} = \softmax_j{(s^k_{i,j})}
\end{gather} Where $\ma{w}^Q_k$ and $\ma{w}^K_k$ are parameters, $\ve{h}$ are hidden states, $d_k$ is the head dimension. These two learning objectives share the same attention matrix, therefore, task-relevant information is implicitly ensured to be preserved in these binary codes.

More specifically, to leverage the multi-head mechanism, they allow each head to represent one and only one bit. By increasing the number of heads to $K$, they obtain attention scores from $K$ different semantic aspects. During the inference stage, codes are generated by binarizing these scores, \begin{gather}
  \ve{c}_i = [c^1_i,\cdots,c^K_i] \in \B^K \\
  c^k_i = \sign{(s^k_{i,i})} \nonumber
\end{gather} During the training stage, they approximate Hamming similarity by computing the cosine similarity, with one of its inputs binarized first. \begin{equation}
  s(i,j) = \cos{(\ve{s}_{i,i},\ve{c}_j)}
\end{equation}

Apart from this similarity function, they also propose a novel loss by carefully selecting instances and eliminating potential positives and negatives. They fine-tune the entire model using both the downstream task loss and the contrastive hashing loss, i.e., $\loss = \loss_{\text{task}} + \beta \cdot \loss_{\text{contrastive}}$. Experiments show that they can reproduce the original performance on an extremely tiny model using only these 24-bit codes as inputs. Therefore, they claim that these codes preserve all the necessary task-relevant information.\par

\section{Proposed Methods}

Our model attempts to learn parsing and hashing simultaneously with a single structured contrastive hashing loss. In other words, we try to introduce the binary representation to output layers and eliminate the need for the $\loss_{\text{task}}$ above. To achieve this, we first extend the CKY module to support binary labels (\S\ref{sec:bit_cky}). Then, we replace the cosine similarity with a newly defined similarity function (\S\ref{sec:similarity}) based on span marginal probabilities, because it contains not only label information but also structural information. After that, we analyze several commonly used contrastive losses, and propose a new one (\S\ref{sec:instance_selection}) to mitigate the geometric center issue as shown in Figure~\ref{fig:center}. After training, we build code vocabulary by mapping binary codes back to their most frequently coinciding labels.\par

\subsection{Constituency Parsing with Bits}
\label{sec:bit_cky}

We decompose tree scores as the sum of constituent scores as well, but with discrete labels $y_i$ replaced with binary codes $\ve{c}_i\in \B^{K}$. \begin{gather}
  g(\ve{t}) = \sum_{\langle l_i,r_i,\ve{c}_i\rangle\in\ve{t}}{g(l_i,r_i,\ve{c}_i)} \\
  g(l_i,r_i,\ve{c}_i) = \sum_{k=1}^{K}{g_k(l_i,r_i,c_i^k)}
\end{gather} Where $g_k(l_i,r_i,c_i^k)$ represents the span score with the $k$-th bit position assigned as value $c_i^k$. We additionally assume that the bits are independent of each other, so we simply add their scores together to obtain the span score $g(l_i,r_i,\ve{c}_i)$.\par

Following \citet{wang-etal-2023-24}, we maintain the one-head-one-bit design and also utilize attention scores for hashing. Furthermore, since we attempt to eliminate $\loss_{\text{task}}$, we do not need to compute the final outputs of the transformer layer. Therefore, we only retain the query $\ma{w}_k^Q$ and key $\ma{w}_k^K$ to calculate the span score for getting $+1$ in the $k$-th bit position by using the token hidden states of the left and right span boundary $\ve{h}_{l_i}$ and $\ve{h}_{r_i}$. For the $-1$ case, we simply leave its score as 0. \begin{align}
  g_k(l_i,r_i,+1) &= \dfrac{(\ma{w}_k^Q \ve{h}_{l_i})^{\top}(\ma{w}_k^K \ve{h}_{r_i})}{\sqrt{d_k}} \\
  g_k(l_i,r_i,-1) &= 0
\end{align}

With these definitions, we can extend the CKY module to the bit-level and calculate the conditional probability and partition function using Equation~\ref{eq:prob} as usual. Additionally, the bit-level marginal probability is defined as below. \begin{equation}
  \mu_k{(l_i,r_i,c_i^k)} = \dfrac{\partial\,\log{Z}}{\partial\,g_k(l_i,r_i,c_i^k)} \label{eq:margin}
\end{equation}

\subsection{Contrastive Hashing with Structures}
\label{sec:similarity}

\citet{wang-etal-2023-24} emphasize that the key to their loss function is first hashing one of its inputs as codes and then calculating the similarity between continuous scores and the discrete codes. We define our similarity function in a similar way. Since we can straightforwardly obtain the span marginal probabilities with Equation~\ref{eq:margin}, we then binarize scores into codes towards the sides with the higher span marginal probabilities. \begin{gather}
  \ve{c}_i = [c_i^1,\dots,c_i^K]\in \B^K \label{eq:binarize} \\
  c_i^k = \begin{cases}
    +1 & \mu_k(l_i,r_i,+1)>\mu_k(l_i,r_i,-1)  \\
    -1 & \text{otherwise} \nonumber
  \end{cases}
\end{gather} 

Naturally, we define the similarity function between two spans as the marginal probability of selecting $i$-th span while assigning $\ve{c}_j$ as its code. \begin{equation}
  s(i,j) = \dfrac{1}{K}\sum_{k=1}^{K}{\mu_k{(l_i,r_i,c_j^k)}} 
\end{equation}

As we mentioned above (\S\ref{sec:parsing}), we use marginal probabilities to define the similarity function because they reflect the joint probability of both selecting the specified span and assigning the specified label to it. If a span is unlikely to be selected as a phrase, then both $\mu_k(l_i, r_i, +1)$ and $\mu_k(l_i, r_i, -1)$ will be close to zero. Thus, the model learns structural and label information simultaneously. By leveraging this similarity function, we extend contrastive hashing to structured contrastive hashing. This approach eliminates the label embedding from the output layer, as the hashing layer returns labels in binary format now.\par

Moreover, for a sentence with $n$ tokens, the total number of spans is $(n^2+n)/2$, and bit-level CKY returns marginal probabilities for them all. However, using them for contrastive hashing leads to an intractable time complexity of $\mathcal{O}{(n^4)}$. In practice, we select only spans from the target trees, reducing the number of spans to $2n-1$ to maintain the time complexity at $\mathcal{O}{(n^2)}$. This is another reason why we prefer marginal probability.\par

\subsection{Instance Selection}
\label{sec:instance_selection}

Following the contrastive learning framework \cite{gao-etal-2021-simcse, wang-etal-2023-24}, we feed sentences into the neural network twice to obtain two semantically identical but slightly augmented representations. In this way, we get two different marginal probabilities for each span. We calculate contrastive losses by comparing each span across views and average them as the batch loss (\S\ref{sec:training}). Note that each batch contains spans from multiple sentences, so our contrastive hashing also compares spans across different sentences. For clarity, we omit subscripts in the following equations when there is no ambiguity.\par

As we mentioned above (\S\ref{sec:contrastive}), the fundamental concept of contrastive learning is pulling together positives and pushing apart negatives. The most commonly used objective function is defined as, \begin{equation}
  \loss_{\text{self}} = -\log{\dfrac{\exp{s(i,i)}}{\sum_{j\in\ca{n}\cup\ca{p}}{\exp{s(i,j)}}}}
\end{equation} $\ca{n}=\set{j \mid y_{j} \neq y_i }$ and $\ca{p}=\set{j \mid y_{j} = y_i }$ stands for the negative and positive sets, respectively. We additionally define $\ca{s}=\set{i}$ as the set that contains span $i$ as its only entry. It is obvious that $\ca{s}\subseteq \ca{p}$ always holds.\par

In addition, the $\log{\sum{\exp}}$ operator is commonly considered a differentiable approximation of the $\max$ operator. Therefore, by slightly tweaking the equation, we reinterpret it as below. \begin{align}
  \loss_{\text{self}} &= \log{\sum_{j}{\exp{s(i,j)}}} - s(i,i) \nonumber \\
  &\approx \negmath{\max_{j\in\ca{n}\cup\ca{p}}{s(i,j)}} - \posmath{s(i,i)} \label{eq:self_loss}
\end{align}

Moreover, \citet{NEURIPS2020_d89a66c7} also proposed a loss function for supervised settings where multiple positives are present. By applying the same tricks, we can rewrite the loss function as follows. \begin{align}
  \loss_{\text{sup}} &= -\dfrac{1}{\abs{\ca{p}}} \sum_{p\in\ca{p}}{\log{\dfrac{\exp{s(i,p)}}{\sum_{j\in\ca{n}\cup\ca{p}}{\exp{s(i,j)}}}}} \nonumber \\
  &= \log{\sum_{j}{\exp{s(i,j)}}}  - \dfrac{1}{\abs{\ca{p}}}\sum_{p}{s(i,p)} \nonumber \\
  &\approx \negmath{\max_{j\in\ca{n}\cup\ca{p}}{s(i,j)}} - \posmath{\mean_{j\in\ca{p}}{s(i,j)}} \label{eq:sup_loss}
\end{align}

\citet{wang-etal-2023-24} have also proposed a contrastive loss function for hashing. They claim that identical tokens may even not contain identical information due to different contexts. Therefore, they treat $\ca{p}$ as potential false positives and negatives, and replace it with $\ca{s}$ in both terms. \begin{align}
  \loss_{\text{hash}} &= -\log{\dfrac{\exp{s(i,i)}}{\sum_{j\in\ca{n}\cup\ca{s}}{\exp{s(i,j)}}}} \nonumber \\
  &\approx \negmath{\max_{j\in\ca{n}\cup\ca{s}}{s(i,j)}} - \posmath{s(i,i)} \label{eq:hash_loss}
\end{align}

By unifying them all in a common format, we can observe that their main differences lie in the instance selection strategies. Both $\loss_{\text{self}}$ and $\loss_{\text{sup}}$ pull instances towards the geometric center of their positive instances in \postext{the second term}. The only difference is that $\loss_{\text{self}}$ assumes there is only one positive instance, making the geometric center merely $i$-th span itself, whereas $\loss_{\text{sup}}$ has access to the ground-truth labels and thus can obtain a more specific center. On the other hand, $\loss_{\text{hash}}$ differs from $\loss_{\text{self}}$ in \negtext{the first term}, as it suggests that dividing instances solely based on ground-truth labels may also introduce false positives and negatives. Therefore, $\loss_{\text{hash}}$ excludes potential false positives from this term. Additionally, it is noteworthy that among these three losses, all \negtext{first terms} employ $\max$, while all \postext{second terms} use $\mean$. Intuitively speaking, the $\max$ operator pulls towards the most likely true positive instance, while $\mean$ operator pulls towards the geometric center of all positive instances. Although it is hard to determine whether instances in $\ca{p}$ are false positives or not, what we can be certain of is that there is at least one true positive instance, since $\ca{s}\subseteq\ca{p}$ always holds. Therefore, using the $\max$ operator to pull towards the most probable one is a more effective approach. \begin{align}
  \loss_{\text{max}} &\approx \negmath{\max_{j\in\ca{n}\cup\ca{s}}{s(i,j)}} - \posmath{\max_{j\in\ca{p}}{s(i,j)}} \nonumber \\
   &\approx \log{\sum_{j\in \ca{n}\cup\ca{s}}{\exp{s(i,j)}}} \nonumber\\ &\quad\quad\quad- \log{\sum_{p\in \ca{p}}{\exp{s(i,p)}}} \nonumber \\
   &= -\log{\dfrac{\sum_{p\in \ca{p}}{\exp{s(i,p)}}}{\sum_{j\in \ca{n}\cup\ca{s}}{\exp{s(i,j)}}}} \label{eq:any_loss}
\end{align}

\subsection{Architecture}
\label{sec:arch}

As shown in Figure~\ref{fig:arch}, our model consists of a pre-trained language model, an attention hash layer, and a bit-level CKY. The attention hash layer derives from the transformer layer, but it preserves only the query and key components for calculating span score $g(l_i,r_i,\ve{c}_i)$. All the other components, such as layer normalization \cite{ba2016layer} and feed-forward layers, are removed, as there is no need to calculate hidden states.\par

Compared with existing constituency parsers \cite{kitaev-klein-2018-constituency,ijcai2020p0560}, the output layer of our parser does not include label embedding matrices. Instead, it utilizes the attention hash layer and a bit-level CKY to predict the binary representation of labels. Consequently, the number of parameters of this part changes from $\abs{\ca{y}} \times d$ to two $K \times \lceil \dfrac{d}{K} \rceil \times d$.\par

\subsection{Training and Inference}
\label{sec:training}

During the training stage, we feed sentences into the model twice to obtain marginal probabilities $\mu^1$ and $\mu^2$ of the two views, and binarize spans on the target trees $\prd{\ve{t}}{}^{1}=\set{\langle l_i,r_i,\ve{c}_i^{1}\rangle}_{i=1}^{2n-1}$ and $\prd{\ve{t}}{}^{2}=\set{\langle l_i,r_i,\ve{c}_i^{2}\rangle}_{i=1}^{2n-1}$ using Equation~\ref{eq:binarize}, respectively. Besides, we use the ground-truth labels $y_i$ to divide $\ca{n}$ and $\ca{p}$. After that, we calculate the contrastive hashing loss for each span with Equation~\ref{eq:any_loss} and average them as the batch loss.\begin{equation}
  \loss = \mean_{1\le i < 2n}{\left(\loss(i,\mu^{1},\prd{\ve{t}}{}^2) + \loss(i,\mu^{2},\prd{\ve{t}}{}^1)\right)} \label{eq:batch_loss}
\end{equation}

After training, we switch the model to evaluation mode, i.e., turning off dropouts \cite{JMLR:v15:srivastava14a}, and then feed all the training sentences into the model again. During this pass, we count the frequency of each pair of binary code and its corresponding ground-truth label, i.e., $f{(\ve{c},y)}$. Then, we can reconstruct a code vocabulary to map codes back to their most frequently coinciding labels. \begin{gather}
  y_{\ve{c}} \gets \argmax_{y\in \ca{Y}}{f{(\ve{c},y)}}
\end{gather}

During the inference stage, we search the most probable tree from all valid trees using the Cocke-Kasami-Younger (CKY) algorithm \cite{kasami1966efficient}. Our bit-level parsers return span boundaries and their binary codes, we translate them back to labels by using the Equation above.\par

\begin{table}[!t]
    \centering
    \adjustbox{width=\columnwidth}{
    \begin{tabular}{lccc}
    \toprule
    \multicolumn{1}{c}{\multirow{2}{*}{\textsc{Model}}} & \multicolumn{2}{c}{\textsc{Ptb}} & \multicolumn{1}{c}{\textsc{Ctb}} \\ 
    & \textsc{\small Bert} & \textsc{\small XLNet} & \textsc{\small Bert} \\
    \midrule
    \citet{kitaev-etal-2019-multilingual}${}^\cky$ & 95.59 & - & 91.75 \\
    \citet{zhou-zhao-2019-head}${}^\cky$ & 95.84 & 96.33 & 92.18 \\
    \citet{mrini-etal-2020-rethinking}${}^\cky$ & - & 96.38 & 92.64 \\
    \citet{ijcai2020p0560}${}^\cky$ & 95.69 & - & 92.27 \\ 
    \citet{NEURIPS2020_f7177163}${}^\transition$ & 95.79 & 96.34 & \sota{93.59} \\
    \citet{tian-etal-2020-improving}${}^\cky$ & 95.86 & \sotb{96.40} &  \sotb{92.66} \\
    \citet{nguyen-etal-2021-conditional}${}^\seq$ & 95.70 & - & - \\
    \citet{xin-etal-2021-n}${}^\cky$ & 95.92 & - & 92.50 \\
    \citet{cui-etal-2022-investigating}${}^\cky$ & 95.92 & - & 92.31 \\
    \citet{yang-tu-2022-bottom}${}^\seq$ & \sotb{96.01} & - & - \\
    \citet{yang-tu-2023-dont}${}^\seq$ & \sota{96.04} & \sota{96.48} & 92.41 \\
    \midrule
    Ours (6 bits)${}^{\cky}$ & 94.81 & 95.70 & 91.45 \\
    Ours (8 bits)${}^{\cky}$ & 95.95 & 96.34 & 91.99 \\
    Ours (10 bits)${}^{\cky}$ & \sota{96.03} & 96.37 & \sotb{92.25} \\
    Ours (12 bits)${}^{\cky}$ & 96.00 & 96.36 & \sota{92.33} \\
    Ours (14 bits)${}^{\cky}$ & \sotb{96.02} & \sota{96.43} & 92.06 \\
    Ours (16 bits)${}^{\cky}$ & 95.98 & \sotb{96.40} & 92.18 \\
    \bottomrule
    \end{tabular}
    }
    \caption{The constituency parsing results. The \sota{bold numbers} and the \sotb{underlined numbers} indicate the best and the second-best performance of each column. $\cky\transition\seq$ stands for the graph-based, transition-based, and sequence-to-sequence models, respectively.}
    \label{tab:exp_parsing}
\end{table}

\begin{table}[!t]
    \centering
    \adjustbox{width=\columnwidth}{
    \begin{tabular}{lccc}
    \toprule
    \multicolumn{1}{c}{\multirow{2}{*}{\textsc{Model}}} & \multicolumn{1}{c}{\textsc{Ace'4}} & \multicolumn{1}{c}{\textsc{Ace'5}} & \multicolumn{1}{c}{\textsc{Genia}} \\ 
    & \textsc{\small Bert} & \textsc{\small Bert} & \textsc{\small Bert} \\
    \midrule
    \citet{wang-etal-2020-pyramid}${}^\spann$ & 86.28 & 84.66 & 79.19 \\
    \citet{wang-etal-2021-nested}${}^\seqlabeling$ & 86.06 & 84.71 & 78.67 \\
    \citet{Xu_Huang_Feng_Hu_2021}${}^\seqlabeling$ & 86.30 & 85.40 & 79.60 \\
    \citet{Fu_Tan_Chen_Huang_Huang_2021}${}^\cky$ & 86.60 & 85.40 & 78.20 \\
    \citet{yu-etal-2020-named}${}^\spann$ & 86.70 & 85.40 & \sotb{80.50} \\
    \citet{shen-etal-2021-locate}${}^\spann$ & 87.41 & 86.67 & \sota{80.54} \\
    \citet{ijcai2021p0542}${}^\spann$ & 87.26 & \sotb{87.05} & 80.44 \\
    \citet{lou-etal-2022-nested}${}^\cky$ & \sotb{87.90} & 86.91 & 78.44 \\
    \citet{zhu-li-2022-boundary}${}^\spann$ & \sota{87.98} & \sota{87.15} & - \\
    \citet{yang-tu-2022-bottom}${}^\seq$ & 86.94 & 85.53 & 78.16 \\
    \midrule
    Ours (4 bits)${}^{\cky}$ & 85.81 & 83.37 & 73.54 \\
    Ours (6 bits)${}^{\cky}$ & 87.39 & 85.23 & 78.57 \\
    Ours (8 bits)${}^{\cky}$ & \sota{87.93} & \sota{85.90} & \sota{78.79} \\
    Ours (10 bits)${}^{\cky}$ & \sotb{87.87} & \sotb{85.75} & \sotb{78.72} \\
    Ours (12 bits)${}^{\cky}$ & 87.52 & 85.26 & 78.40 \\
    \bottomrule
    \end{tabular}
    }
    \caption{The nested named entity recognition results. $\cky\spann\seqlabeling\seq$ stands for graph-based, span-based, sequential labeling, and sequence-to-sequence models, respectively.}
    \label{tab:exp_nest}
\end{table}

\section{Experiments}
\label{sec:exp}

\subsection{Settings}
\label{sec:settings}

We validate the effectiveness of our model on various structured prediction tasks, i.e., constituency parsing and nested named entity recognition tasks. Dataset statistics can be found in Appendix~\ref{sec:data_stats}.\par

For the constituency parsing task, we conduct experiments on the datasets PTB \cite{marcus-etal-1993-building} and CTB5.1 \cite{XUE_XIA_CHIOU_PALMER_2005}. We
transform the original trees into those of Chomsky
normal form and adopt left binarization with \texttt{NLTK} \cite{bird-loper-2004-nltk}. We study model performance by employing pre-trained language models with checkpoints \texttt{bert-large-cased} \cite{devlin-etal-2019-bert} and \texttt{xlnet-large-cased} \cite{NEURIPS2019_dc6a7e65} for PTB, and \texttt{bert-base-chinese} for CTB.\par
 
For nested named entity recognition task, we use datasets ACE2004 \cite{doddington-etal-2004-automatic}, ACE2005 \cite{walker2006ace}, and GENIA \cite{kim2003genia}. We follow the data splitting of \citet{shibuya-hovy-2020-nested}. Nested named entity recognition, as \citet{Fu_Tan_Chen_Huang_Huang_2021} claimed, can be considered as a partially observed constituency parsing task. Therefore, we add a dummy span \texttt{TOP} as the top span to each sentence to ensure all the observed spans form a valid parsing tree, and apply the same transformation and binarization on it. We use the checkpoint \texttt{bert-large-cased} on ACE2004 and ACE2005, and for the GENIA dataset we use \texttt{dmis-lab/biobert-large-cased-v1.1} \cite{Lee_2019} as the pre-trained language model.\par

We utilize the deep learning framework \texttt{PyTorch} \cite{NEURIPS2019_bdbca288} to implement our models and download pre-trained lanugage checkpoints from \texttt{huggingface/transformers} \cite{wolf-etal-2020-transformers}.\par

To keep the training of contrastive hashing stable, we collect sentences until the total number of tokens in each batch reaches 1024. We employ Adam optimizer \cite{kingma2017adam,loshchilov2018decoupled}, and the total number of training steps of constituency parsing and nested named entity recognition are \num{50000} and \num{20000}, and the warm-up are \num{4000} and \num{2000} steps, respectively. To provide harder negatives by augmenting inputs, we also randomly mask a portion of tokens as \texttt{[MASK]}.\par

Experiments are all conducted on a single NVIDIA Tesla V100 graphics card, the total training wall time is around 3 hours and 1 hour, respectively. All experiments are run with two different random seeds and the reported numbers in the following tables are their averages.\par

\begin{table}[!t]
    \centering
    \begin{tabular}{cccc}
    \toprule
    \negtext{\textsc{Neg}} & \postext{\textsc{Pos}} & \textsc{Loss} & \textsc{Ptb} \\
    \midrule
    \multirow{3}{*}{$\max_{\ca{n}\cup\ca{p}}$} & ${}_{\ca{s}}$ & $\loss_{\text{self}}$ & 81.08 \\
     & $\mean_{\ca{p}}$ & $\loss_{\text{sup}}$ & 95.58 \\
     & $\max_{\ca{p}}$ & - & 95.75 \\
    \midrule
    \multirow{3}{*}{$\max_{\ca{n}\cup\ca{s}}$} & ${}_{\ca{s}}$ & $\loss_{\text{hash}}$ & 94.26 \\
     & $\mean_{\ca{p}}$ & - & \sotb{95.88} \\
     & $\max_{\ca{p}}$ & $\loss_{\text{max}}$ & \sota{96.03} \\
    \bottomrule
    \end{tabular}
    \caption{Ablation study of instance selection strategies in constituency parsing experiments. The columns \textsc{Neg} and \textsc{Pos} display the selection strategies for negatives and positives, respectively. \textsc{Loss} shows this combination corresponds to which loss definition.}
    \label{tab:ablation}
\end{table}

\subsection{Main Results}
\label{sec:main_results}

Our model consistently achieves competitive performance on various structured prediction tasks and datasets, as presented in Table~\ref{tab:exp_parsing} and Table~\ref{tab:exp_nest}.\par

For constituency parsing, our models reach the peak with around 12 bits. Continuously increasing the number of bits does not further improve performance, on the contrary, it leads to a slight decline. We attribute this to the disproportionately large hashing space, as the amount of information carried by each task and dataset is fixed. For example, assigning $K$ bits to a task with only $K$ labels leads to an extreme case. Models in this case tend to produce the most trivial bit-level one-hot representation, making them nothing different from traditional static embedding models. On the contrary, decreasing the number of bits to fewer than 8 bits is also harmful, due to the insufficient representation capability. Besides, our models outperform almost all previous graph-based methods that rely on maximizing the log-likelihoods of target trees \cite{kitaev-etal-2019-multilingual,mrini-etal-2020-rethinking,ijcai2020p0560,xin-etal-2021-n,cui-etal-2022-investigating}. Therefore, we claim that leveraging contrastive learning is beneficial to representation learning.\par

For nested named entity recognition, all datasets show the best performance at the 8-bit settings, and decreasing to fewer than 6 bits also results in insufficient representing capability. Similarly, our methods outperform the previous sequential labeling methods \cite{shibuya-hovy-2020-nested,wang-etal-2021-nested,xin-etal-2021-n} and graph-based methods \cite{Fu_Tan_Chen_Huang_Huang_2021,lou-etal-2022-nested}. In addition, some other papers \cite{yu-etal-2020-named,shen-etal-2021-locate,ijcai2021p0542,zhu-li-2022-boundary} propose to straightforwardly enumerate all spans and directly train the model to classify them. These methods currently show the best performance, our method can also achieve comparable results to them.\par

\begin{table}[!t]
  \centering
  \adjustbox{width=0.86\columnwidth}{
  \begin{tabular}{ccc}
  \toprule
  \textsc{Label} & \textsc{Code} & \textsc{Coverage (\%)} \\
  \midrule
  \multirow{2}{*}{\texttt{S }} & \texttt{110110001000} & $56.48$ \\
   & \texttt{110100001101} & $42.80$ \\
  \midrule
  \multirow{2}{*}{\texttt{S'}} & \texttt{101100001000} & $59.69$ \\
   & \texttt{110110101001} & $38.36$ \\
  \midrule
  \texttt{NP } & \texttt{010101001101} & $99.38$ \\
  \midrule
  \texttt{NP'} & \texttt{010001010100} & $98.70$ \\
  \midrule
  \texttt{VP } & \texttt{100010100111} & $99.52$ \\
  \midrule
  \texttt{VP'} & \texttt{001010010111} & $98.23$ \\
  \midrule
  \multirow{2}{*}{\texttt{ADJP }} & \texttt{100100000111} & $66.22$ \\
   & \texttt{000011010110} & $29.08$ \\
  \midrule
  \texttt{ADJP'} & \texttt{000000010110} & $93.34$ \\
  \midrule
  \multirow{2}{*}{\texttt{ADVP }} & \texttt{101100010111} & $84.53$ \\
   & \texttt{000101100111} & $11.87$ \\
  \midrule
  \multirow{2}{*}{\texttt{ADVP'}} & \texttt{001100010110} & $52.03$ \\
   & \texttt{000101110110} & $37.00$ \\
  \bottomrule
  \end{tabular}}
  \caption{Example of the hashing results on the constituency parsing task. The \textsc{Label} column shows the labels and their corresponding incomplete labels, which are introduced during the Chomsky normal form transformation. The \textsc{Code} and \textsc{Coverage} columns display binary codes and their frequency proportions among all possible codes under each label. For instance, label $\texttt{S'}$ is supposed to be assigned to an incomplete span within a larger $\texttt{S}$ span, and 59.69\% of $\texttt{S'}$ labels are translated from the code \texttt{101100001000}.}
  \label{tab:code}
\end{table}

\begin{figure*}[t]
\centering
\adjustbox{width=0.98\textwidth}{
\begin{tikzpicture}
\Tree [.\code{S}{D88} [.\code{NP}{54D} She ]
   [.\code{VP}{8A7} ate
       [.\code{NP}{54D} [.\code{NP}{54D} the pumpkin ]
           [.\code{SBAR}{D2B} [.\code{WHNP}{71C} that ]
                 [.\code{S}{D88} [.\code{NP}{54D} Luna ]
                    [.\code{VP}{8A7} smashed ] ] ] ] ] ]
\end{tikzpicture}
\begin{tikzpicture}
\Tree [.\code{S}{D06} [.\code{NP}{54D} The quick brown fox ]
   [.\code{VP}{8A7} jumps
       [.\code{PP}{56A} over
       [.\code{NP}{54D} the lazy dog ] ] ] ]
\end{tikzpicture}}
\caption{Examples of the hashing and constituency parsing results. The hexadecimal numbers in the brackets indicate the generated binary codes, and the span labels are translated from them.}
\label{fig:tree}
\end{figure*}
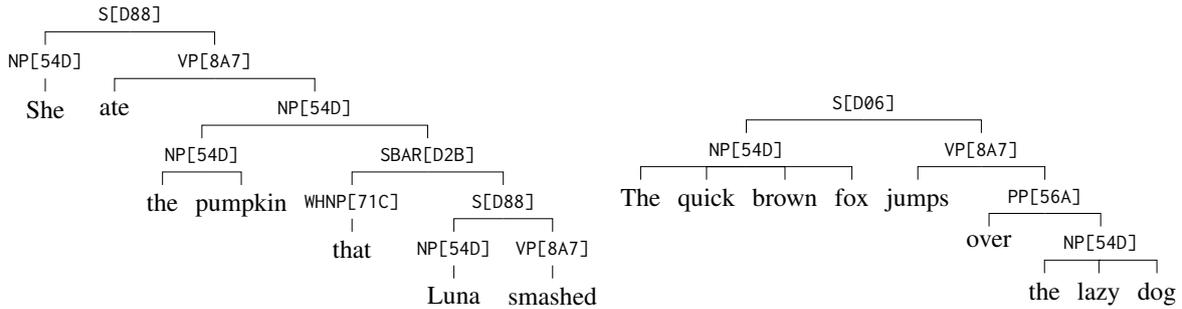

\subsection{Ablation Studies}
\label{sec:ablation}

We enumerate all combinations of the contrastive learning objective functions in Table~\ref{tab:ablation} to study the influence of instance selection strategies.\par

We note that selecting ${}_{\ca{n}\cup\ca{s}}$ is always superior to selecting ${}_{\ca{n}\cup\ca{p}}$ in the \negtext{negative term}, but using ${}_{\ca{s}}$ is inferior to using ${}_{\ca{p}}$ in the \postext{positive term}. This leads to the conclusion that adding positive instances $\ca{p}$ to the negative term is harmful, but it turns out to be beneficial when added to the positive term. This seems contradictory to the conclusion of \citet{wang-etal-2023-24}. However, this discrepancy arises because the training mechanisms are different. Their model relies on downstream task loss to implicitly learn structural information, whereas our model is designed to learn it explicitly. Therefore, we claim for the setting that aims at straightforwardly hashing both the label and structural information using a single loss, our $\loss_{\text{max}}$ is the best choice.\par

Besides, \citet{ouchi-etal-2020-instance,ouchi-etal-2021-instance} appear to be the most relevant publications to our research. They also attempt to directly apply contrastive learning methods to structured prediction tasks. However, they pull instances towards the geometric centers, and their results show that this does not lead to noteworthy improvements than conventional graph-based models. This conclusion is consistent with ours, i.e., $\max_{\ca{p}}$ is better than $\mean_{\ca{p}}$. We hypothesize that calculating the geometric center largely erases the contextual information of each instance. Pulling towards this center is not essentially different from pulling towards a static label embedding. Therefore, their approach loses the advantage of contrastive learning and trivially falls back to the conventional static embedding classification. On the contrary, our $\max_{\ca{p}}$ primarily targets the closest instance instead, therefore, it can fully leverage the capability of contrastive learning and distribute instances more uniformly, and allows each label to associate with multiple codes.\par

\subsection{Case Studies}
\label{sec:case_studies}

We show some frequent labels and their associated codes in Table~\ref{tab:code}. We only display the codes that occupy the top $90\%$ frequency of each label. It can be observed that some labels have two codes with roughly the same frequency, while others are predominantly represented by a single code. However, even in these cases, none reach $100\%$ coverage. This result further demonstrates that the term $\max_{\ca{p}}$ imposes only a weak supervision signal and does not force all instances towards the geometric centers, thereby avoiding false positives and allowing a more uniform distribution of instances, rather than crowding at a few specific points.\par

In Figure~\ref{fig:tree}, we display some parsing examples, using the same sentences as \citet{wang-etal-2023-24}. Our parser takes discrete tokens as inputs and returns binary labels, while theirs, conversely, receives binary codes of tokens and parses them into trees with discrete labels. Our codes preserve not only static by also contextual information of each label, for instance, the two \texttt{S} labels above are translated from two different codes. Therefore, our methods can also be considered as implicitly clustering methods, since these two codes refer to two different sub-labels of label \texttt{S}. This phenomenon is unlikely to be observed in models suffering from the geometric center issue.\par

\section{Related Work}

The ease of training has allowed contiguous representation \cite{doi:10.1126/science.1127647} to achieve great success in applications such as masked language models \cite{devlin-etal-2019-bert,liu2020roberta} and causal language models \cite{peters-etal-2018-deep,lewis-etal-2020-bart,openai2023gpt4,touvron2023llama}. Different from them, auto-encoder \cite{JMLR:v11:vincent10a,kingma2022autoencoding,pmlr-v32-rezende14} and contrastive learning \cite{He_2020_CVPR,gao-etal-2021-simcse} do not classify instances into specific classes, but rather reconstruct the input or determine whether a given pair belongs to the same class. The advantage of these approaches is that they do not require introducing embedding matrices, making representation learning more effective. However, because embeddings are not introduced, continuous vectors are hard to be discretized to specific classes and need to train an additional classifier for such purpose.\par

Learning discrete representation has also been widely focused on since the early era of deep learning \cite{pmlr-v5-salakhutdinov09a,pmlr-v15-courville11a,pmlr-v32-mnih14,pmlr-v48-mnihb16}, as it is a necessary component of many downstream applications. For example, languages are inherently discrete, and mapping discrete tokens to continuous vectors and back is the foremost step for modern natural language processing. Additionally, many applications involving the control and interpretation of intermediate variables also require interleaving non-differentiable layers. However, the key challenge is that backpropagating gradients through discrete representation is generally intractable, making them hard to train.\par

The most straightforward solution is the straight-through estimator \cite{DBLP:journals/corr/BengioLC13,yin2018understanding}, which simply copies gradients across non-differentiable layers, treating the Jacobian matrix as an identity matrix. SPIGOT \cite{peng-etal-2018-backpropagating,mihaylova-etal-2020-understanding} further introduces a structured projection to reduce gradient errors. Building upon this previous work, VQ-VAE \cite{NIPS2017_7a98af17,NEURIPS2019_5f8e2fa1} allows interleaving a discrete layer with continuous inputs mapped as discrete codes. These returned informative codes preserve necessary information for downstream tasks, and at the same time, provide an interface for controlling and interpreting the hidden states in intermediate layers of neural networks.\par

Another line of research turns to relaxation techniques, as they approximate discrete representation with continuous relaxation to mitigate the intractability issue. The Gumbel-softmax estimator \cite{jang2017categorical} was first proposed by using the Gumbel-max trick \cite{NIPS2014_309fee4e,maddison2017the} to provide a differentiable approximation of the $\argmax$ operation. The marginal approaches \cite{DBLP:journals/corr/FriesenD16a,kim2017structured,liu-lapata-2018-learning,eisner-2016-inside} even further relax the discreteness restriction, not limiting themselves to selecting only a single discrete token, but instead aggregating information from all discrete tokens according to softmax distribution.\par

Different from all of them, our binary representation lies between continuous and discrete representations. On the one hand, the easy-to-train property of continuous representation allows us to avoid designing complex gradient estimators while also benefiting from state-of-the-art representation learning methods, i.e., contrastive learning. On the other hand, the easy-to-discretize property allows us to instantly transform these continuous representations into discrete representations, thus making the imposition of various task-relevant structures tractable and efficient. Besides, in other potential applications such as machine translation and language models, using binary representation removes the need for a large softmax, thus, reduces memory consumption and accelerates training and inference. Additionally, the final step of re-interpreting binary codes as labels or tokens allows the model to 
preserve much more contextual information within the codes. Consequently, this implicit clustering capability may potentially address issues like polysemy through discovering subclasses (\S\ref{sec:case_studies}).\par

\section{Conclusions}

In this paper, we demonstrated that introducing the binary representation to the output side is also feasible. Binary representation inherits the advantages of easy training from contiguous representation and the structure modeling capability from discrete representation. Therefore, binary representation can be considered as a novel representation that further bridges the gap between the continuous nature of deep learning and the discrete intrinsic property of natural languages. We achieved this by extending previous contrastive hashing into structured contrastive hashing. Specifically, we designed a bit-level CKY, defined a similarity function based on marginal probability, and proposed a novel contrastive hashing loss to mitigate the geometric center issue. Experiments show that our methods perform remarkably on various structured prediction tasks by using only a few bits. Our model also demonstrates a certain degree of implicit clustering capability in discovering subclasses. Most importantly, although we primarily focus on structured prediction tasks, our method can be easily applied to other natural language processing tasks by equipping the corresponding structures.\par

\section*{Limitations}

Our contrastive hashing method yields informative binary representations. Nonetheless, information theory suggests that labels with different frequencies carry varying amounts of information. Therefore, compressing all continuous label representation into codes with fixed number of bits is not the optimal solution, and it may even lead to the problem of high-frequency labels being overly concentrated in a narrow area. We remain to solve this issue by extending our method to a variant number of bits in future work.\par

\bibliography{references}

\appendix

\section{Dataset Statistics}
\label{sec:data_stats}

\begin{table}[th]
  \adjustbox{width=\columnwidth}{
  \begin{tabular}{cccccc}
  \toprule
  \textsc{Dataset} & \textsc{Train} & \textsc{Dev} & \textsc{Test} & \textsc{Label} & \textsc{Cnf} \\
  \midrule
  \textsc{Ptb} & \num{39832} & \num{1700} & \num{2416} & 26 & 130 \\
  \textsc{Ctb} & \num{18104} & \num{352} & \num{348} & 26 & 133 \\
  \midrule
  \textsc{Ace'4} & \num{6198} & \num{742} & \num{809} & 8 & 22 \\
  \textsc{Ace'5} & \num{7285} & \num{968} & \num{1058} & 8 & 39 \\
  \textsc{Genia} & \num{15022} & \num{1669} & \num{1855} & 5 & 21 \\
  \bottomrule
  \end{tabular}}
  \caption{Statistics of these five datasets. Columns \textsc{Train}, \textsc{Dev}, and \textsc{Test} show the number of sentences, \textsc{Label} and \textsc{Cnf} display the number of labels before and after Chomsky normal form transformation.}
  \label{tab:stats}
\end{table}

\end{document}